# MaterialFigBENCH: benchmark dataset with figures for evaluating college-level materials science problem-solving abilities of multimodal large language models

Michiko Yoshitake [a]*, Yuta Suzuki [b], Ryo Igarashi [a], Yoshitaka Ushiku [a], and Keisuke Nagato [c]

[a]OMRON SINIC X, Tokyo, Japan; [b]Department of Applied Physics, Osaka Univ, Suita, Japan; [c]Department of Mechanical Engineering, Univ. Tokyo, Tokyo, Japan

Michiko Yoshitake, OMRON SINIC X, Nagase Hongo Building 3F, 5-24-5 Hongo, Bunkyo-ku, Tokyo, Japan 113-0033: michiko.yoshitake@sinicx.com

# MaterialFigBENCH: benchmark dataset with figures for evaluating college-level materials science problem-solving abilities of multimodal large language models

We present MaterialFigBench, a benchmark dataset designed to evaluate the ability of multimodal large language models (LLMs) to solve university-level materials science problems that require accurate interpretation of figures. Unlike existing benchmarks that primarily rely on textual representations, MaterialFigBench focuses on problems in which figures such as phase diagrams, stress–strain curves, Arrhenius plots, and microstructural schematics are indispensable for deriving correct answers. The dataset consists of 137 free-response problems paired with the corresponding answer (or answer range) and associated figure(s), based on standard university-level materials science textbooks, covering a broad range of topics including crystal structures, mechanical properties, diffusion, phase diagrams, phase transformations, and electronic properties of materials. To address unavoidable ambiguity when reading numerical values from images, expert-defined answer ranges are provided where appropriate.

We evaluate several state-of-the-art multimodal LLMs, including ChatGPT and GPT-series models accessed via OpenAI APIs, and analyze their performance across problem categories and model versions. The results reveal that, although overall accuracy improves with model updates, current LLMs still struggle with genuine visual understanding and quantitative interpretation of materials science figures. In a considerable number of cases, correct answers are obtained by relying on memorized domain knowledge rather than on reading the provided images. MaterialFigBench highlights persistent weaknesses in visual reasoning, numerical precision, and significant-digit handling, while also identifying problem types for which performance has improved. This benchmark provides a systematic and domain-specific foundation for advancing multimodal reasoning capabilities in materials science and for guiding the development of future LLMs with stronger figure-based understanding.

# 1. Introduction

After the rapid development of generative large language models (LLMs) such as ChatGPT, Gemini, and Claud 3, the utilization of generative LLMs in the materials science field has been increasingly explored [1-5]. With the emergence of applications of LLMs in materials science, various benchmark datasets have been developed. Previously, we developed a benchmark dataset focused on materials science, composed of text-based questions and answers taken from textbook exercises [6]. A considerable number of benchmark datasets of different types have been constructed. Since information in the materials science field is inherently multimodal, benchmark datasets corresponding to multimodal information have also developed.

There are many multimodalities in materials science, such as molecular structures, crystal structures, various spectra, microscopy images, and phase diagrams. Among them, molecular structures and crystal structures are used as visual images in practical materials science research, but in machine learning they are normally treated not as images but as structured textual data. Molecular structures are often represented in SMILES format, and a benchmark for "Molecular Structure–Text Understanding", MolTextQA has been constructed based on SMILES strings [7]. Crystal structures are described using crystallographic information files (CIFs). Information on crystal structures is usually extracted partially depending on specific research needs, in forms such as lattice constants, coordination numbers, and Voronoi diagrams. In LLM4Mat-Bench [8], a benchmark for evaluating the performance of LLMs in predicting the properties of crystalline materials, CIF files were converted to textual descriptions of crystal structures using Robocrystallographer [9].

As for spectral data, mostly they are recorded as arrays of numbers, which can be treated as vectors. Then, from a viewpoint of input, those data, molecular structures, crystal structures and spectral data are textual. There are other specific benchmarks in material science for textual input such as for synthesis using atomic layer deposition [10] and for materials science tools (solve problems with using tools such as pymatgen) [11].

For data such as microscopy images and phase diagrams, images are typically provided directly as inputs to multimodal LLMs. MMMU [12] provided a benchmark dataset for image input across massive multi-discipline fields, of which 26 % were categorized as Tech & Engineering. However, datasets specific to materials science are limited both in number and diversity. MaCBench [13] provided diverse visual inputs, such as laboratory images, band structures, crystal structures, and atomic force microscopy images, paired with multiple-choice questions. There is also a benchmark focusing on highly specific visual inputs, namely density of state (DOS) images, for LLMs to analyze and interpret electronic structures [14]. In this work, we constructed a benchmark dataset specialized for materials science that incorporates wide variety of image types commonly used in university-level textbooks.

## 2. Benchmark dataset

### 1)  *Selection of problems*

Previously we constructed a text-only benchmark dataset for materials science and evaluated the performance of large language models in solving materials science problems using this dataset, MaterialBENCH [6]. For the earlier benchmark, two university-level materials science textbooks were selected [15,16] that cover a broad spectrum of materials science topics. Textbooks specializing in specific materials, such as *polymers*, or specific phenomena, such as *magnetism*, were excluded. The process of textbook selection was described in detail in ref. 6. The problems in the previous benchmark dataset cover a wide array of topics, including defects, diffusion, fracture, corrosion and various material properties, and span a broad range of materials such as metals, semiconductors, ceramics, polymers, and composites. At that time, problems related mainly phase diagrams and phase transitions were excluded because they require figure inputs. In the present work, we focus on problems which require accurate figure recognition and interpretation.

Problems are selected such that the accompanying figure(s) are indispensable for solving them. Care was taken to exclude problems that include figures but can nevertheless be solved using text alone. Only problems for which correct answers are explicitly provided by the textbook authors were selected.

## 2) *Modification of Problem Texts*

This dataset consists of problem-answer (or answer range) pairs. All problems are composed of textual descriptions and figure(s), require reading or interpreting figures to answer, and are of free-response type rather than multiple-choice. To avoid copyright issues, we manually modified the problem texts and figures while preserving the essential physical and conceptual content of the original problems.

Another type of modification involves dividing problems that contain multiple sub-questions, typically labeled (a), (b), etc., into separate problems. For example, Problem 9-18 in the original textbook— "A 30 wt% Sn–70 wt% Pb alloy is heated to a temperature within the α+liquid phase region. If the mass fraction of each phase is 0.5, estimate: (a) The temperature of the alloy; (b) The compositions of the two phases" — was divided into Problem #61 and #62 in the benchmark dataset. In contrast, when multiple questions are strongly interdependent, they were retained as a single problem. For instance, in Problem #18, the original problem text was modified to: "In the figure of the uploaded EXA_5-5.png file is shown a plot of the logarithm (to the base 10) of the diffusion coefficient versus reciprocal of absolute temperature, for the diffusion of metal-A in metal-B. Determine values for the activation energy Q [kJ/mol] and the preexponential D0 [m2/s]. **Answer the two values separated by a comma.**" In this case, additional clarifying text (shown in bold in the above example) was added, and the problem wase not divided.

When multiple answers are expected for a single question, phases such as "give all phases" or "answer the concentration(s)" were inserted into the problem sentences to ensure that generative models recognize that more than one answer may be correct. In some cases, the original problem texts already implied multiple correct answers, as in Problem #37: "*The figure of the uploaded EXA_9-2.png file is a phase diagram for a hypothetical metal MA and MB*. For a 40 wt% *MB*–60 wt% *MA* alloy at 150 C, **what is (are) the composition(s) of the phase(s)**?", noted as bold. In this example, the phrasing in italics was further refined in accordance with the figure modifications described below.

### 3)     *Modification of Problem Figures*

All the figures were redrawn by one of the authors based on the corresponding textbook figures, so that the physical meanings remain unchanged, while ensuring that the figures are not identical to those in the textbooks. In general, materials names in the figures were replaced by symbolic labels so that the LLMs cannot rely on memorized knowledge of specific materials without interpreting the provided figures. In most problems involving phase diagrams, acutual element names were replaced with hypothetical symbols; for example, the "Cu-Ag phase diagram" was converted to an "MA-MB phase diagram". Figure filenames were also modified to ensure that figures corresponding to each problem are easily distinguishable, regardless of the number of figures provided.

In accordance with the figure modifications, the corresponding problem texts were also revised. In addition to changes necessitated by figure modification, further clarifications— such as explicit unit specifications— were added, following the same approach used in MaterialBENCH [6].

*Examples of Figure Modifications*

Problem #27 (originally Problem 7.25 in the textbook) is used as an illustrative example. The original and modified figures are shown in Fig. 1 (a) and Fig. 1(b), respectively. The original problem sentence reads: "If it is assumed that the plot in Figure 7.15 is for noncold-worked brass, determine the grain size (d) of the alloy in Figure 7.19; assume its composition is the same as the alloy in Figure 7.15." Figure 7.15 corresponds to Fig. 1(a-1). Figure 7.19 contains three plots used throughout the textbook, two of which were modified for this problem and correspond to Fig. 1(a-2) and Fig. 1(a-3).

The following modifications were applied. The slope of the straight line in Fig. 1(a-1) was altered in the modified figure Fig. 1 (b-1), while the intersection value with the pink dotted line—used to answer the question—was preserved. The plot for 1040 steel in Fig. 1 (a-2) was shifted downward and relabeled, while the intersection value of B (originally brass) at zero percent cold work was preserved in Fig. 1(b-2). Similarly, the plot for 1040 steel in Fig. 1 (a-3) was shifted and relabeled, while the plot for B was retained in Fig. 1(b-3). This third plot is not required to solve the problem but was intentionally included to evaluate the LLMs' ability to identify the relevant figure. Figure names were also changed accordingly (e.g., Figure 7.15 -> 7-25-1). Based on these changes, the problem text was revised to "If it is assumed that the plot in the figure of the uploaded 7-25-1.png file is for noncold-worked metal B, determine the

grain size (d [mm]) of the metal B referring the figures of the uploaded 7-25-2.png and 7-25-3.png files."

### 4)     Settings of correct answer range

Problem answers fall into two primary categories: numerical values and textual responses, such as yes/no answers, crystallographic indices, or phase names. For textual answers, ambiguity is minimal, and therefore no answer range was defined. For example, responses such as "possible" and "yes" are treated as equivalent.

When numerical answers are obtained through calculations using values explicitly stated in figures, the expected answer is exact, and no answer range was applied unless issues related to numerical precision or significant digits arose. However, when numerical values must be read directly from images, unavoidable deviations can occur, even for human experts. In such cases, a reasonable answer range was defined instead of a single value.

For problems requiring a single numerical answer derived from a figure, one answer range was assigned. For problems requiring two numerical answers, two independent answer ranges were defined. These ranges were determined by experts, accounting for realistic uncertainties in reading abscissa and ordinate values.

Figure 2(a) illustrates the answer-range determination for Problem #18: "In the figure of the uploaded EXA_5-5.png file is shown a plot of the logarithm (to the base 10) of the diffusion coefficient versus reciprocal of absolute temperature, for the diffusion of metal-A in metal-B. Determine values for the activation energy Q [kJ/mol] and the preexponential D0 [m2/s]. Answer two values by connecting comma.". To compute the activation energy, the slope of the straight line must be evaluated by reading two points from the plot. While uncertainty in the abscissa is negligible, the ordinate values show small deviations. Red lines in Fig. 2(a) support reading the values of abscissa and ordinate of the two positions. The value of $\log_{10}$(ordinate) corresponding to abscissa 0.8 can range from -12.7 to -12.8, and that corresponding to abscissa 1.1 can range from -15.7 to -15.8. These uncertainties lead to an activation energy range of 185.7–198.5 kJ/mol and a pre-exponential factor range of $0.9 \times 10^{-5}$–$3.9 \times 10^{-5}$ m²/s. Model responses within these ranges were considered correct.

Another example is Problem #62 (originally Problem 9-18), which involves estimating phase compositions from a phase diagram, as shown in Fig. 2(b): "The figure of the uploaded 9-18.png file is a phase diagram for a hypothetical metal MA and MB. A 30 wt% MB–70 wt% MA alloy is heated to a temperature within the α+liquid phase region. If the mass fraction of each phase is 0.5, estimate the compositions [wt% MB] of the two phases." Lines and characters in red and blue are added for the explanations of answer range setting (not in the provided image to LLMs). In this example, under the condition "mass fraction of each phase is 0.5" in 30 wt% MB–70 wt% MA alloy, the temperature that the distance from 30 wt% MB–70 wt% MA line to phase boundary line for $\alpha$ is equal to that for liquid should be identified (Problem #61) at first, which is shown as a sold blue line in the figure. Then the compositions of the intersections between the blue line and the two phase-boundary lines should be identified. Based on uncertainties in reading composition values, the acceptable ranges were set to 15–17 wt% MB for the α phase and 41–44 wt% MB for the liquid phase.

5) *Benchmark Dataset Summary*

The resulting benchmark dataset consists of 137 problem–answer (answer-range) pairs, constructed using the procedures described above. All answers provided by the textbook authors were verified and, where necessary, adjusted by an expert to ensure consistency with the modified problems. The problems can be broadly categorized as follows: (1)

crystal orientation and indexing, (2) stress–strain curves and hardness, (3) phase diagrams of hypothetical alloys, (4) Fe–C phase diagrams, (5) Fe–C phase transformations and microstructures, (6) phase diagrams of hypothetical oxide systems, and (7) electrical conductivity of semiconductors.

The selected problems require not only accurate figure interpretation but also solid domain knowledge, logical reasoning, quantitative analysis, and the ability to understand complex materials science concepts. The dataset is designed to represent a comprehensive range of challenging university-level materials science problems. For reference, we provide the original problem numbers as they appear in the textbooks. For each problem, the corresponding correct answer or answer range is provided. The dataset is publicly available on HuggingFace [17].

## 3. Evaluation of LLMs with benchmark dataset

The benchmark dataset was submitted to multiple generative LLMs—ChatGPT-4o, ChatGPT-o1, ChatGPT-o3-mini, ChatGPT-o3-mini-high, ChatGPT-5, ChatGPT-5-auto and ChatGPT-5-thinking. We also evaluated GPT-4o, GPT-o1 and GPT-5 via OpenAI API using the dataset. For all models evaluated here, figure file(s) could be attached as part of the input.

For ChatGPT, only one response for one problem was collected, because responses are generated interactively one by one. For the OpenAI API experiments, only GPT-4o, GPT-o1 and GPT-5 supported figure attachment at the time of the study; therefore, we evaluated these models via the API. GPT-4o allowed up to 10 responses per input, whereas GPT-o1 and GPT-5 were limited to a maximum of 8 responses.

Because we observed that minor figure modifications— such as changes in color, line thickness, or font style in axes and labels— did not lead to differences in LLM responses, we avoided increasing the number of problems by creating near-duplicate items with similar figures that require identical interpretations and logic. This design choice makes straightforward statistical analysis infeasible. Instead, in this section we focus on identifying differences and similarities among models based on observed error patterns, and we discuss underlying causes of these errors in the next section.

1) *Answer Correctness Judgement*

When no answer range was defined (i.e. for textual answers or for numerical values explicitly stated in the images), only an exact answer was accepted. Even when an exact numerical answer was expected, significant-digits handling can affect correctness. For problems in which a single numerical value is obtained by calculation using values written in a figure (image), the response should report the appropriate number of significant digits consistent with the figure. Some LLMs produced values with fewer significant digits, which may reflect an attempt to increase apparent correctness by reducing precision. When the significant digits were not appropriate for problems in which values can be calculated directly from the quantities shown in the figure, the response was regarded as incorrect.

When a response is a single numerical value and correct range is defined as min to max, the response is regarded as correct if min $\leqq$ value $\leqq$ max. For example, in Problem #15, the interplanar spacing for (110) plane is calculated by reading $2\theta$ angle from the X-ray diffraction pattern provided as a figure. Suppose the model response is 0.203 [nm], which corresponds to interpreting $2\theta$ as 44.7°. If $2\theta$ can reasonable be read as 43–47°, the correct answer range is 0.1934-0.2104 [nm] (min = 0.1934, max = 0.2104). Because 0.203 satisfies $0.1943 \leq 0.203 \leq 0.2104$, the response is judged as correct.

When a model response is represented as a range ($min^{value}$ to $max^{value}$), the response is regarded as incorrect if the reported range only touches the boundary (e.g., $max^{value}$ = min or $min^{value}$ = max). For example (Problem #84), if the correct answer range is 250–

275 and the model responds 220–250 (max$^{value}$ = min), the response is judged as incorrect. When there is an overlap between the reported range and the correct range, the response is judged as correct. For example, if the correct answer range is 250–275 and the model responds with 270–300, the response is judged as correct.

A small number of problems allow more than one correct answer (e.g., two different compositions satisfy the required condition). In such cases, we modified the problem sentences so that LLMs were instructed to provide all valid answers, as described in Section 2-2). For these problems, responses were judged correct only when all valid answers were given and each fell within the correct answer range.

Another type of multi-answer problem involves physically paired quantities, such as treatment temperature and duration of treatment, or activation energy and a pre-exponential factor. For these responses, we judged the response as correct only when all reported quantities were within their corresponding correct answer ranges.

All judgements of response were performed by an expert in this study.

## 2) *Overall Results*

Overall results are summarized in Table 1 for ChatGPT and in Table 2 for OpenAI API models, together with the dates on which the problem statements were submitted. For ChatGPT, each problem was submitted once. For OpenAI API, multiple responses were collected per problem: ten for GPT-4o and eight for GPT-o1 and GPT-5 (the maximum allowed for these models).

The accuracies listed in Table 1 are 0.328, 0.438, 0.343, 0.336, 0.409, 0.482, and 0.555 for ChatGPT-4o, ChatGPT-o1, ChatGPT-o3-mini, ChatGPT-o3-mini-high, ChatGPT-5, ChatGPT-5-auto, and GPTGPT-5-thinking, respectively. At the time of writing, ChatGPT-o1, ChatGPT-o3-mini, and ChatGPT-o3-mini-high are not available. When problems were submitted using ChatGPT-5, there was no option to select sub-models.

Table 1 shows that the later models, ChatGPT-o3-mini and ChatGPT-o3-mini-high achieved substantially lower scores than ChatGPT-o1. Notably, the score of ChatGPT-o1 (0.438) exceeds that of ChatGPT-5, released in August 2025.

It is unclear that the accuracy differences between ChatGPT-4o and ChatGPT-o3-mini or ChatGPT-o3-mini-high are meaningful. Although not shown, around in May we evaluated ChatGPT-o3 and ChatGPT-4o-mini-high on Problems #1–#35 (categories #2 and #3), for which other models mostly failed to provide correct answers. ChatGPT-o3 produced 10 correct answers out of 35, and ChatGPT-4o-mini-high produced 11 out of 35. For comparison, the number of correct answers for Problem #1–#35 were 7 (ChatGPT-4o), 10 (ChatGPT-o1), 6 (ChatGPT-o3-mini), and 6 (ChatGPT-o3-mini-high). Thus, the performance of ChatGPT-o3 and ChatGPT-4o-mini-high on this subset is comparable to that ChatGPT-o1. This is disappointing, given that ChatGPT-o4-mini-high emphasizes visual reasoning. It should also be noted that even among ChatGPT-5 variants, performance on these 35 problems remained similar: 9 (ChatGPT-5), 12 (ChatGPT-5-auto), and 10 (ChatGPT-5-thinking).

Table 2 summarizes the API-based results for GPT-4o (API-4o), GPT-o1 (API-o1), and GPT-5 (API-5). To obtain an accuracy value, we divided the number of correct responses by the number of sampled responses (10 for GPT-4o and 8 for GPT-o1 and GPT-5) and then averaged over all 137 problems. As expected, accuracy increased with model updates. For reference, the corresponding ChatGPT accuracies are also listed in Table 2.

Unexpectedly, the API-based accuracies CPT-4o and GPT-o1 (0.257 and 0.350) were lower than those of ChatGPT-4o (0.331) and ChatGPT-o1 (0.431), even though the API responses were collected later. Although only one response per problem was collected for ChatGPT, these differences appear substantial. For version 5, only GPT-5 is available via the API, whereas multiple ChatGPT-5 variants exist and have changed over time. In contrast to 4o and o1, the API-based accuracy for GPT-5 was higher than that of any ChatGPT-5 variant. When using ChatGPT-5, deep-thinking was effectively skipped because no reasoning process was displayed in deep-thinking mode. Therefore, it is understandable that ChatGPT-5 showed the lowest score among ChatGPT-5, ChatGPT-5-auto and ChatGPT-5-thinking. The API GPT-5 appears most compatible to ChatGPT-5-thinking, and there was no substantial difference between their accuracy scores.

*3) Evaluation of ChatGPT Models*

The patterns of correct and incorrect answers are shown in Fig. 3, where pink denotes problems answered correctly and grey denotes incorrect answers. The red lines separate earlier models from ChatGPT-5 variants.

Across models, some problems consistently show high correctness while others show low correctness. Only for category #4 (Fe-X phase diagram problems) were LLM answers correct with high probability. Correct answers were especially rare for the category #1 (crystal orientation and indices) and category #2 (strain-stress curves and hardness) across all models, and for category #7 (electrical conductivity of semiconductors) for earlier models.

For category #4, despite replacing carbon (C) with a hypothetical element X, both ChatGPT and API-based GPT models often inferred X as carbon and obtained correct answers by using memorized solubility limits rather than reading the attached figures. We verified this effect by attaching figures in which solubility values were removed. This behavior helps explain why LLMs sometimes produced correct answers even without an image attachment, as discussed in Section 4.1.

On the other hand, when we intentionally showed slightly different eutectic compositions in the figures, both ChatGPT and API-based GPT models correctly used the modified solubility-limit values in the figures, as confirmed by additional tests with modified figures.

As seen in Fig.3, accuracy generally increased from 4o, o1 to 5, consistent with model upgrades. However, o3-mini and o3-mini-high, which were released after o1, exhibited lower overall accuracy than o1 in this study. Because only one response per problem was obtained for each ChatGPT model, it is difficult to discuss problem-level differences in detail. Nevertheless, category-level accuracy patterns are similar across models, and no clear category-specific improvement trend was observed across versions.

### 4) Evaluation of GPTs Models (OpenAI-API)

The patterns of correct and incorrect answers are shown in Fig. 4, where pink denotes correct and grey denotes incorrect. For reference, the ChatGPT pattern from Fig. 3 is also shown.

Figure 4 indicates that (1) GPT-o1 achieved higher accuracy than GPT-4o, and (2) the categories with high and low scores closely resemble those observed for ChatGPT: high for category #4 and low for categories #1, #2, #3, #5, #6 and #7.

As noted above for the category #4 (Fe-C phase diagram problem), despite replacing carbon with a hypothetical element X, GPT models also often inferred X as carbon and produced correct answers using memorized solubility limits rather than the attached figures. We also found that GPT models sometimes answered even without attaching figures when the problems text stated "refer to the attached figure". In such cases, GPT models inferred X as carbon, used learned values, and calculated answers, leading to a high apparent accuracy.

Fig. 5 shows the distribution of accuracy rates for GPT-4o, GPT-o1, and GPT-5. The abscissa denotes the per-problem accuracy rate, and the ordinate denotes the number of problems with the corresponding accuracy. Consistent with the overall accuracies, the number of problems with accuracy 0 decreases in the order 4o > o1 > 5, whereas the number with accuracy 1 shows the opposite trend. A characteristic feature across all models is the U-shaped distribution, with peaks near 0 and 1 and relatively few problems in between. This suggests that correctness is driven by problem characteristics rather than random variation.

To further examine differences among API models, we compared problems that were answered correctly in all trials by one model but incorrectly in all trials by another, as summarized in Table 3.

For example, problems that were always incorrect with GPT-4o but always correct with GPT-o1 were #87, #109, #113, and #125. Problems that were always incorrect with GPT-4o but always correct with GPT-5 were #40, #45, #46, #50, #54, #57, #59, #65, #96, #106, #115, #117, and #125; only #125 overlapped between these two sets. Problems that were always incorrect with GPT-o1 but always correct with GPT-5 were #50, #54, #57, #96, #98#, 106, and #117, which largely overlaps with the GPT-4o vs. GPT-5 comparison.

In contrast, Problem #21 was always correct with GPT-4o but always incorrect with the higher models GPT-o1 and GPT-5. Due to significant-digit handling: GPT-4o responded with three significant digits, whereas GPT-o1 and GTP-5 responded with two. There were no problems that were always correct with GPT-o1 but always incorrect with GPT-5. Notably, the behavior observed for Problem #21 in the API setting is consistent with the ChatGPT setting (using the ChatGPT-5-thinking variant). Similarly, problems that were always incorrect with GPT-4o but always correct with

GPT-o1 (#109, #113, #125) largely overlapped with the corresponding ChatGPT comparisons.

*5) Comparison Between ChatGPT and API-Based GPT*

For the comparison below, we use responses from ChatGPT-5-thinking variant as representative of ChatGPT-5.

For the 4o version, the number of problems for which ChatGPT was correct and API-GPT as always correct was 14, and the number for which ChatGPT was incorrect and API-GPT was always incorrect was 55, yielding an agreement on 69 problems. The number of problems for which ChatGPT was correct but API-GPT was always incorrect was 8 (#23, #39, #40, #44, #48, #82, #104, #136), while the reverse case (ChatGPT incorrect, API-GPT always correct) occurred for 1 problem (#86), yielding complete disagreement on 9 problems.

For the o1 version, the number of problems for which both ChatGPT and API-GPT were always correct was 25 (10 of which overlapped with the 4o agreement set), and the number of problems for which both were always incorrect was 47 (37 overlapping with the 4o agreement set), yielding agreement on 72 problems**.** There are 8 problems for which ChatGPT was correct but API-GPT was always incorrect (#23, #28, #85, #89, #98, #102, #126, #136), and there are no problems for the reverse case, yielding complete disagreement on 8 problems.

For version 5, there were 49 problems for which ChatGPT was correct and API-GPT was always correct, and 30 problems for which ChatGPT was incorrect and API-GPT was always incorrect, yielding agreement on 79 problems. In the opposite direction, there was 1 problem for which ChatGPT was correct but API-GPT was always incorrect (#19), and 2 problems for which ChatGPT was incorrect but API-GPT was always correct (#96, #98), yielding complete disagreement on 3 problems.

Overall, with the model update, agreement increased and complete disagreement decreased, indicating that the discrepancies between ChatGPT and API-based GPT outputs narrowed.

4.Response Analysis

*1) Response Without Figure(s)*

Surprisingly, generative LLMs often produced responses even when problem statements were input without attaching figure file(s), and these responses were frequently correct. In some models or for some problems, LLMs requested that the figure be attached; however, in most cases, responses were generated without explicitly asking for figures. For example, all ChatGPT variants produced the correct answer without using the figure for Problem #71: "The figure of the uploaded 9-51.png file is a partial phase diagram for iron and hypothetical element X. Consider 2.5 kg of austenite containing 0.65 wt% X, cooled to below 727C. How many kilograms each of pearlite and the proeutectoid phase form?". ChatGPTs correctly inferred that X corresponds to carbon and used known solubility values—0.022 for the solubility limit of carbon in ferrite, and 0.76 for the eutectic composition—from memorized knowledge rather than form the provided figure, thereby producing the correct answer. Because extensive information on the Fe-C phase diagram is widely available online, we avoided Fe-C related problems when checking API-based responses, as discussed below.

Problems used for this verification were #3 (directional index) "Answer the directional index (four-index system) for the direction shown in blue arrow in the figure of the uploaded EXA_3-9.png file.", #15 (X-ray diffraction pattern) "The figure of uploaded 3-64.png file shows an X-ray diffraction pattern for a hypothetical FCC metal taken using a diffractometer and monochromatic x-radiation having a wavelength of 0.1542 nm; each diffraction peak on the pattern has been indexed. Calculate the interplanar spacing for (110) plane.", #18 (Arrhenius plot) "In the figure of the uploaded EXA_5-5.png file is shown a plot of the logarithm (to the base 10) of the diffusion coefficient versus reciprocal of absolute temperature, for the diffusion of metal-A in metal-B. Determine values for the activation energy Q [kJ/mol] and the preexponential D0 [m2/s]. Answer two values by connecting comma.", #67 (phase diagram) "The figure of the uploaded 9-45.png file shows the pressure–temperature phase diagram for a hypothetical material. Apply the Gibbs phase rule at the black point; that is, specify the number of degrees of freedom at the point—that is, the number of externally controllable variables that need be specified to completely define the system." #78 (Fe-C-X phase diagram) "The figure of the uploaded 9-66-c.png file is a partial phase diagram for iron–carbon system. Consider a steel alloy containing 93.8 wt% Fe, 6.0 wt% metal-N, and 0.2 wt% C, where metal-N is a hypothetical element. What is the proeutectoid phase when this alloy is cooled to a temperature just below the eutectoid? Refer to the figures of the uploaded 9-66-a.png and 9-66-b.png files, on which M, N, S, T and W represent different hypothetical element added." #110 (histogram) "Assume that the molecular weight distributions shown in the figure of the uploaded EXA_14-1.png file are for a hypothetical polymer, where the molecular weight of the repeat unit is 62.50 g/mol. For this material, compute the number-average molecular weight." A brief summary of these results is provided in Table 4.

*2) Reading of Numerical Value Texts in Figures*

When exact numerical values required for calculations were explicitly given as texts within figures, the models consistently read and used these values correctly. For example, in Problem #136, depending on the uploaded figures—Fig. 6(a) (B8-3) or Fig. (b) (B8-3-a) — the eutectoid carbon composition and the maximum solubility of carbon in ferrite phase are 0.76 and 0.022 in Fig. 6(a), and 0.77 and 0.0218 in Fig. 6 (b), respectively.

When the problem statement "The figure of the uploaded xxx.png file is a partial phase diagram of iron-carbon alloy. A steel having a carbon concentration of 0.65% contains, when cooled from austenite, certain amounts of primary ferrite and pearlite. What are the carbon contents of these two microconstituents?" (xxx is either B8-3 or B8-3-a) was input with Figure 6(a), the response was "Primary ferrite (α): 0.022 wt% C, Pearlite (eutectoid austenite): 0.76 wt% C", whereas with Figure 6(b) the response was "Primary ferrite (α): ~0.022 wt% C (≈0.0218), Pearlite (α + Fe$_3$C microconstituent): ~0.77 wt% C (the eutectoid composition)". These responses clearly indicate that the LLMs correctly recognized the numerical values in the figures and appropriately rounded them according to significant-digits conversions. Therefore, there is little doubt regarding the models' ability to read numerical value texts embedded in figures.

*3) Significant Digits*

Significant-digit handling emerged as a critical issue. For example, in Problem#21— "From the tensile stress–strain behavior for the specimen shown in the figure of the uploaded EXA_6-3.png file, determine the maximum load [N] that can be sustained by a cylindrical specimen having an original diameter of 12.8 mm." —GPT-4o returned values with five significant digits (typically 57,915 N), calculated as 450 [N/mm²] × 128.7 [mm²]. Because such values can be appropriately rounded to three significant digits, these responses were regarded as correct.

In contract, GPT-o1 consistently returned answers with only two significant digits, such as 58 [kN] or 5.8 x 10$^4$ [N], despite using intermediate values with three or more significant digits during calculation. These responses were therefore regarded as incorrect. GPT-5 showed similar behavior, producing answers with only two significant digits even when higher precision was used internally.

A similar trend was observed for ChatGPT variants: ChatGPT-4o produced answers with three significant digits, ChatGPT-o1 with two digits, ChatGPT-5 and ChatGPT-5-auto with three digits, and ChatGPT-5-thinking with two digits. Overall, higher-tier models within the same family tended to report fewer significant digits, possibly to increase apparent correctness without explicitly considering significant-digit requirements.

*4) Accuracy Improvement with Model Updates*

Problems for which accuracy clearly improved with model updates were relatively limited: #3 (crystal index of hexagonal lattice), #45 (Cu-Ag phase diagram: phase identification), #65 (Cu-Ag phase diagram: mass fraction), #68 (Fe-C phase diagram: ferrite fraction), #100 (two graphs, Fe-C system: hardness feasibility), #132 (Ag-Cu phase diagram: liquefaction temperature), and #137 (Arrhenius plot: bandgap as activation energy).

Despite the presence of many alloy phase diagrams beyond Cu-Ag system, no clear accuracy improvements were observed for those systems. This suggests that LLMs may have learned Cu-Ag phase diagram images together with explanatory text, but not other phase diagrams. Consequently, GPT models do not yet demonstrate a general capability to read and interpret phase diagrams. A similar conclusion applies to crystal orientation and plane-reading tasks: although accuracy improved for Problem #3, accuracy for other similar problems remained low.

For Problem #100, which requires a yes/no answer regarding the feasibility of achieving specified hardness and ductility, it is likely that LLMs relied on pretrained domain knowledge rather than image interpretation. This interpretation is supported by the persistently low accuracy for related problems (#27 and #104), which require careful reading and interpretation of multiple figures.

In contrast, performance on Arrhenius-type plots appears to have improved. Not only did Problem #137 show higher accuracy with newer models, but the correct answer rate for Problem #18 increased from 0.3 (GPT-4o) to 0.625 (GPT-o1 and GPT-5). Notably, although GPT-5 achieved a higher overall accuracy, there were still many problems for which GPT-5 performed worse than GPT-o1.

### *5) Problems with Complete Failure or Complete Success*

There were 37 problems for which none of the evaluated models produced correct answers. Among these, eight problems (#4-7, #9-10, #12-13) involved crystal orientation indices. Excessive rounding caused incorrect responses for Problems #42, #55, and #127. Problems involving isothermal transformation diagrams (#92 and #95) also proved particularly challenging.

Two cases are especially noteworthy, First, although the Cu-Ag phase diagram appears to be well learned by LLMs, all models failed on Problem #64, which requires measuring distances on the phase diagram and performing calculations based on those measurements. Second, for Problem #118, which concerns the temperature dependence of intrinsic carrier concentration in Si on a logarithmic scale, all models failed, whereas they succeeded for the corresponding Ge problem using the same image. This discrepancy strongly suggests reliance on memorized material-specific knowledge rather than image interpretation.

Conversely, there were 10 problems for which all models succeeded. Among these, six (#69, #72, #78, #93, #134, #135) involved the Fe-C phase diagram, where complex image interpretation was unnecessary. It is interesting to note the difference on the behavior between Problem #93, and #92 and #95, of which all involve an isothermal transformation diagram. While all models successfully solved Problem #93, they completely failed to solve Problems #92 and #95, which resemble Problem #93 but correct answer involved multiple phases.

Problem #129 provides one the clearest demonstrations of genuine image-based reasoning. The problem asks whether a hypothetical MA–MB alloy can exist with specified solid and liquid compositions. Although all API-based models produced correct answers, the quality of reasoning and explanations improved substantially with model updates, with GPT-5 providing the most explicit and physically consistent explanation.

Other problems that were consistently answered correctly—#20, #36, and #67—require relatively straightforward image reading and appear to be comparatively easy for LLMs. Overall, no misunderstandings of problem text were observed, in contrast to earlier evaluations using the text-only MaterialBENCH dataset (November 2023-May 2024). LLMs generally followed the required logical steps correctly; incorrect answers primarily resulted from failures in reading and interpreting images rather than from misunderstandings of the problem statements.

## 5. Discussion and Implications

This study reveals several important implications for the design of multimodal large language models (LLMs) as well as for the construction of benchmark datasets intended to evaluate visual reasoning in materials science.

First, the results demonstrate that high answer accuracy does not necessarily indicate genuine visual understanding. Across multiple problem categories, LLMs frequently produced correct answers without relying on the provided figures, instead exploiting memorized domain knowledge learned during pretraining. This behavior was particularly evident for canonical materials systems such as the Fe–C phase diagram, where extensive prior knowledge is widely available. Consequently, benchmarks that do not explicitly control for this effect risk overestimating true multimodal reasoning capability. Future benchmark design should therefore prioritize problems in which figure-based reasoning is indispensable and cannot be bypassed through memorization alone.

Second, the observed performance disparities across problem categories highlight fundamental limitations in current multimodal perception. Tasks requiring geometric interpretation—such as crystal orientation indexing, diffraction-peak analysis, or distance measurement in phase diagrams—remain especially challenging. In contrast, tasks involving Arrhenius-type plots showed measurable improvement with model updates, suggesting that progress in visual reasoning may be uneven and representation-dependent. These findings indicate that advances in multimodal LLMs do not uniformly translate across different classes of scientific figures, underscoring the need for fine-grained, domain-specific evaluation.

Third, significant-digit handling emerged as a nontrivial weakness. Several higher-tier models consistently reported answers with reduced numerical precision, even when higher precision was clearly supported by the figures. This behavior suggests that current models may implicitly prioritize semantic correctness over numerical rigor, a tendency that is problematic for scientific and engineering applications. Incorporating explicit training signals or evaluation criteria related to numerical precision and significant-digit conventions may therefore be necessary to improve the reliability of LLMs in quantitative scientific reasoning.

From the perspective of LLM development, these results imply that improving multimodal performance requires more than scaling model size or general visual capabilities. Instead, targeted training on scientifically meaningful visual abstractions— such as phase boundaries, logarithmic slopes, and crystallographic directions—may be essential. Additionally, mechanisms that encourage explicit grounding of answers in visual inputs, rather than latent textual knowledge, could help mitigate shortcut behaviors observed in this study.

From the benchmark-design perspective, MaterialFigBench illustrates the importance of domain-specific benchmarks that reflect authentic problem-solving practices in science and engineering. Unlike generic multimodal benchmarks, materials science problems often require precise quantitative reading, multi-step reasoning, and careful interpretation of specialized figures. Benchmarks that incorporate these characteristics, along with expert-defined answer ranges and controlled figure modifications, provide a more realistic assessment of multimodal reasoning capabilities.

Overall, this work suggests that current multimodal LLMs remain far from robust figure-based understanding in materials science. MaterialFigBench offers a foundation for systematically diagnosing these limitations and tracking progress over time. We expect that such benchmarks will play a critical role in guiding the development of future LLMs that can genuinely reason from scientific figures, rather than merely recall learned textual patterns.

## 6. Conclusions

In this work, we introduced MaterialFigBench, a benchmark dataset designed to evaluate multimodal large language models on university-level materials science problems that require accurate interpretation of scientific figures. Through a systematic evaluation of state-of-the-art LLMs, we demonstrated that current models often achieve correct answers without genuinely regarding the provided figures, instead relying on memorized domain knowledge. While modest improvements were observed for certain figure types, such as Arrhenius plots, substantial limitations persist in tasks that demand precise geometric interpretation, quantitative measurement, and proper handling of significant digits.

These findings indicate that existing multimodal LLMs remain far from robust figure-based reasoning in materials science. MaterialFigBench provides a domain-specific and methodologically grounded framework for diagnosing these shortcomings and for tracking future progress. We expect that this benchmark will contribute to the development of next-generation multimodal models that can reliably ground their reasoning in scientific figures, thereby enabling more trustworthy applications of LLMs in materials science research and education.

References:

[1] Hatakeyama-Sato K, Yamane N, Igarashi Y, Nabae Y, Hayakawa T. Prompt engineering of GPT-4 for chemical research: what can/cannot be done? Science and Technology of Advanced Materials: Methods, 3(1) (2023). https://doi.org/10.1080/27660400.2023.2260300.

[2] Yu S, Ran N, Liu J. Large-language models: The game-changers for materials science research. Artificial Intelligence Chemistry, 2(2) (2024) 100076. https://doi.org/10.1016/j.aichem.2024.100076.

[3] Lei G, Docherty R, Cooper SJ. Materials science in the era of large language models: a perspective. Digital Discovery. 3 (2024) 1257–1272. https://doi.org/10.1039/d4dd00074a

[4] Jiang X, Wang W, Tian S, Wang H, Lookman T, Su Y. Applications of natural language processing and large language models in materials discovery. npj Computational Materials, 11 (2025) 79.

[5] Miret, S., Krishnan, N.M.A. Enabling large language models for real-world materials discovery. Nat Mach Intell, 7, 991–998 (2025). https://doi.org/10.1038/s42256-025-01058-y

[6] Yoshitake M, Suzuki Y, Igarashi R, Ushiku Y, Nagato K. MaterialBENCH: Evaluating college-level materials science problem-solving abilities of large language models. arXiv:2409.03161.

[7] Laghuvarapu S, Lee N, Gao C, Sun J. MolTextQA: A question-answering dataset and benchmark for evaluating multimodal architectures and LLMs on molecular structure–text understanding. Journal of Data-centric Machine Learning Research (2025). https://openreview.net/forum?id=tFfqvKE2J5

[8] Rubungo AN, Li K, Hattrick-Simpers J, Dieng AB. LLM4Mat-bench: benchmarking large language models for materials property prediction. Machine Learning: Science and Technology, 6 (2025) 020501.

[9] Ganose AM, Jain A. Robocrystallographer: automated crystal structure text descriptions and analysis. MRS Communications, 9(3) (2019) 874–881.

[10] Yanguas-Gil A, Dearing MT, Elam JW, Jones JC, Kim S, Mohammad A, Nguyen CT, Sengupta B. Benchmarking large language models for materials synthesis: the case of atomic layer deposition. Journal of Vacuum Science & Technology A, 43 (2025) 032406. https://doi.org/10.1116/6.0004319.

[11] Liu S, Xu J, Ye B, Hu B, Srolovitz DJ, Wen T. MatTools: Benchmarking large language models for materials science tools. arXiv:2505.10852. https://doi.org/10.48550/arXiv.2505.10852.

[12] Yue X, Ni Y, Zhang K, Zheng T, Liu R, Zhang G, Stevens S, Jiang D, Ren W, Sun Y, Wei C, Yu B, Yuan R, Sun R, Yin M, Zheng B, Yang Z, Liu Y, Huang W, Sun H, Su Y, Chen W. MMMU: A massive multi-discipline multimodal understanding and

reasoning benchmark for expert AGI. 2024 IEEE/CVF Conference on Computer Vision and Pattern Recognition (CVPR) IEEE, DOI: 10.1109/CVPR52733.2024.0091, 9556-9567. arXiv:2311.16502.

[13] Alampara N, Mandal I, Khetarpal P, Grover HS, Schilling-Wilhelmi M, Krishnan NMA, Jablonka KM. MaCBench: a multimodal chemistry and materials science benchmark. Proceedings of the 38th Conference on Neural Information Processing Systems (NeurIPS 2024). https://openreview.net/forum?id=Q2PNocDcp6.

[14] Takahara I, Mizoguchi T. Benchmarking multimodal large language models on electronic structure analysis and interpretation. AI for Accelerated Materials Design – NeurIPS Poster, (2025). https://openreview.net/forum?id=LPIhtHdYZF

[15] Hummel RE. Understanding Materials Science. 2nd ed. Springer; 2004.

[16] Callister WD Jr, Rethwisch DG. Materials Science and Engineering: An Introduction. 8th ed. John Wiley & Sons; 2010.

[17] HuggingFace: https://huggingface.co/omron-sinicx

Table 1. Results of the evaluation using the constructed benchmark dataset.

| | Table 1 Results of the evaluation using the constructed benchmark dataset | | | | | | |
|---|---|---|---|---|---|---|---|
| model | ChatGPT-4o | ChatGPT-o1 | ChatGPT-o3-mini | ChatGPT-o3-mini-high | ChatGPT-5 | ChatGPT-5:auto | ChatGPT-5:think |
| accuracy | 0.328 | 0.438 | 0.35 | 0.343 | 0.409 | 0.482 | 0.555 |
| date of experiments | 2025.02.13-2025.02.27 | 2025.02.13-2025.02.27 | 2025.02.13-2025.02.27 | 2025.02.13-2025.02.27 | 2025.08.20-2025.08.21 | 2025.09.18-2025.09.24 | 2025.09.18-2025.09.24 |

Table 2. Results of the evaluation using the constructed benchmark dataset

| Table 2 Results of the evaluation using the constructed benchmark dataset | | | |
|---|---|---|---|
| model | API-4o | API-o1 | API-5 |
| accuracy | 0.264 | 0.37 | 0.581 |
| date of experiments | 2025.04.17-2025.06.05 | 2025.04.24-2025.06.05 | 2025.08.28-2025.09.11 |
| cf. accuracy in corresponding ChatGPT | 0.328 | 0.438 | 5: 0.409<br>5-auto: 0.482<br>5-think: 0.555 |

Table 3. Problems that were answered correctly in all trials by one model but incorrectly in all trials by another models.

| | | all correct | | |
|---|---|---|---|---|
| | | GPT-4o | GPT-o1 | GPT-5 |
| all incorrect | GPT-4o | – | 4 (#87, #109, #113, #125) | 13 (#40, #45, #46, #50, #54, #57, #59, #65, #96, #106, #115, #117, **#125**) |
| | GPT-o1 | 1 (#21) | – | 7 (#50, #54, #57, #96, #98, #106, #117) |
| | GPT-5 | 1 (#21) | none | – |

Table 4. Summary of responses without attaching figures for different models. GPT-5 responded most reasonably way saying it cannot view figures, but other models responded by guessing from memorized knowledge.

|  | #3 | #15 | #18 | #67 | #78 | #110 |
|---|---|---|---|---|---|---|
| 4o | all: cannot view | calculation equations given: we do not actually have the specific value of \\( \\theta \\) from the image data you referenced; Without the specific \\(\\theta\\) value from the diffraction data | 3: cannot view, calculation equations | Assuming the black point indicates a triple point , The "black point" you referenced is likely a point where multiple phases coexist, Without the specifics of the diagram, I'll assume the black point | explaine how to & asuuming ferrite | all: In the absence of the actual data from the uploaded figure (which I cannot view) |
| o1 | 5: cannot view 2: without knowing exactly which start- and end-points define the blue arrow in your figure. | calculation equations only | Typical numerical values from such a plot (often encountered for a substitutional metal diffusing in another metal) are on the order of\n\n• Q ≈ 140\u202fkJ u202fmol⁻¹, \n• D₀ ≈ 2 ×10⁻⁵\u202fm² | all: the black point in the diagram= the triple point | no mention on Figure, but simply answer ferrite | how to: Because the fractional data from "EXA_14-1.png" is not reproduced here, In many textbook examples (including Callister's Example 14.1–type problems, |
| 5 | all: cannot view | 3: cannot view 1: 0.256 nm | all: cannot view | all: cannot view | all: from the 9-66 figures | all: cannot view |

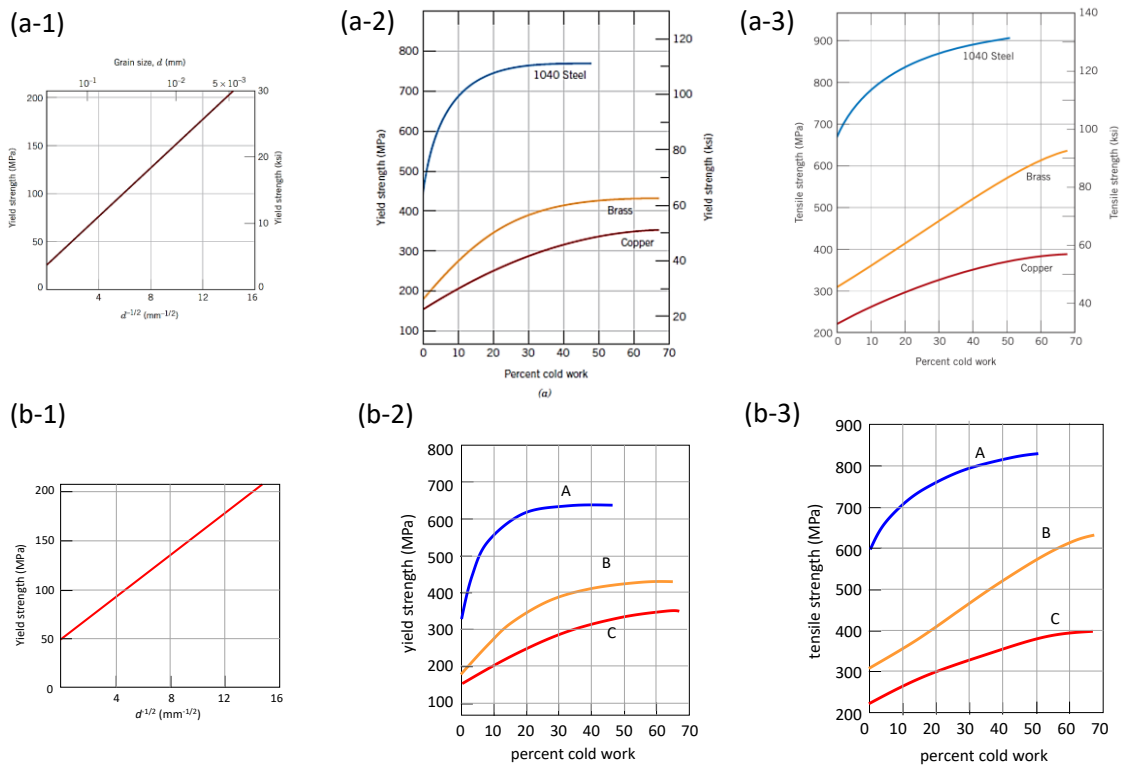

Figure 1. Examples of figure alterations: (a) original ones, (b) corresponding altered ones.

Figure 2. Example of correct answer range setting: (a) scale reading, (b) distance measuring and cross point reading.

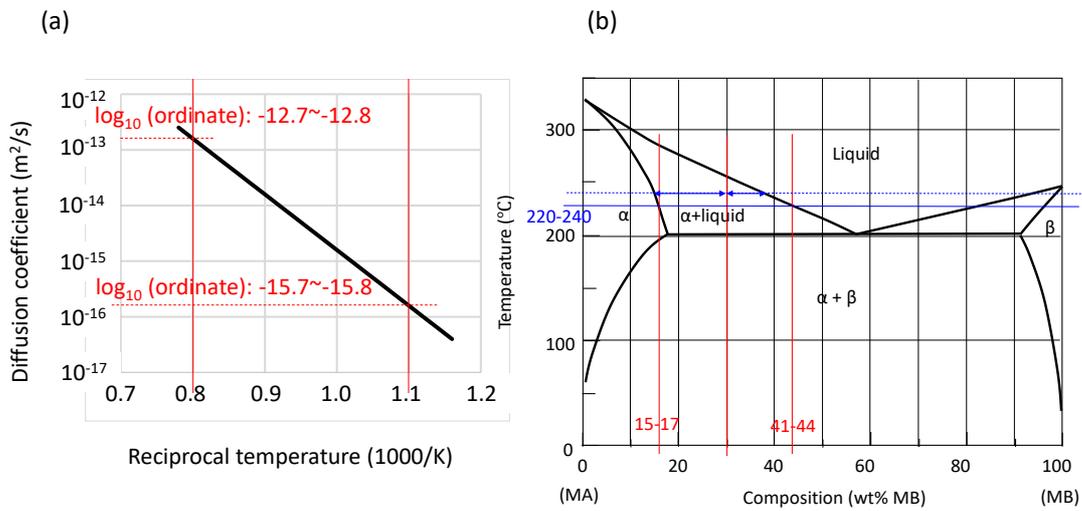

Figure 3. The patterns of correct and incorrect answers from different ChatGPT models. Pink represents correct, grey incorrect. Number denotes categories explained in the text.

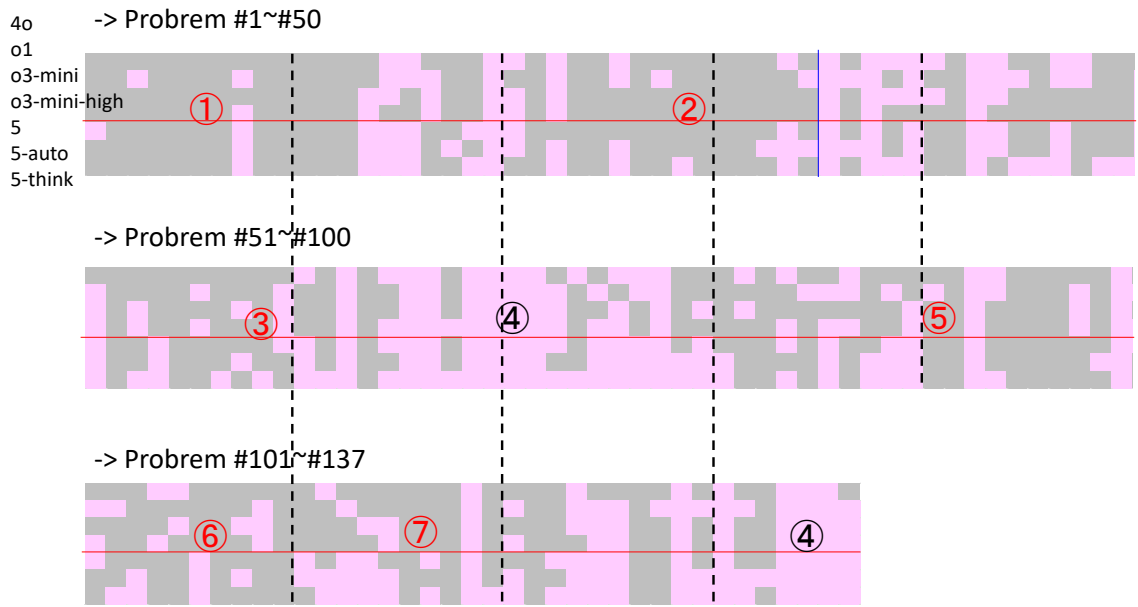

Figure 4. The patterns of correct and incorrect answers from different API-GPT models together with those of ChatGPT models. Pink represents correct, grey incorrect. Number denotes categories explained in the text.

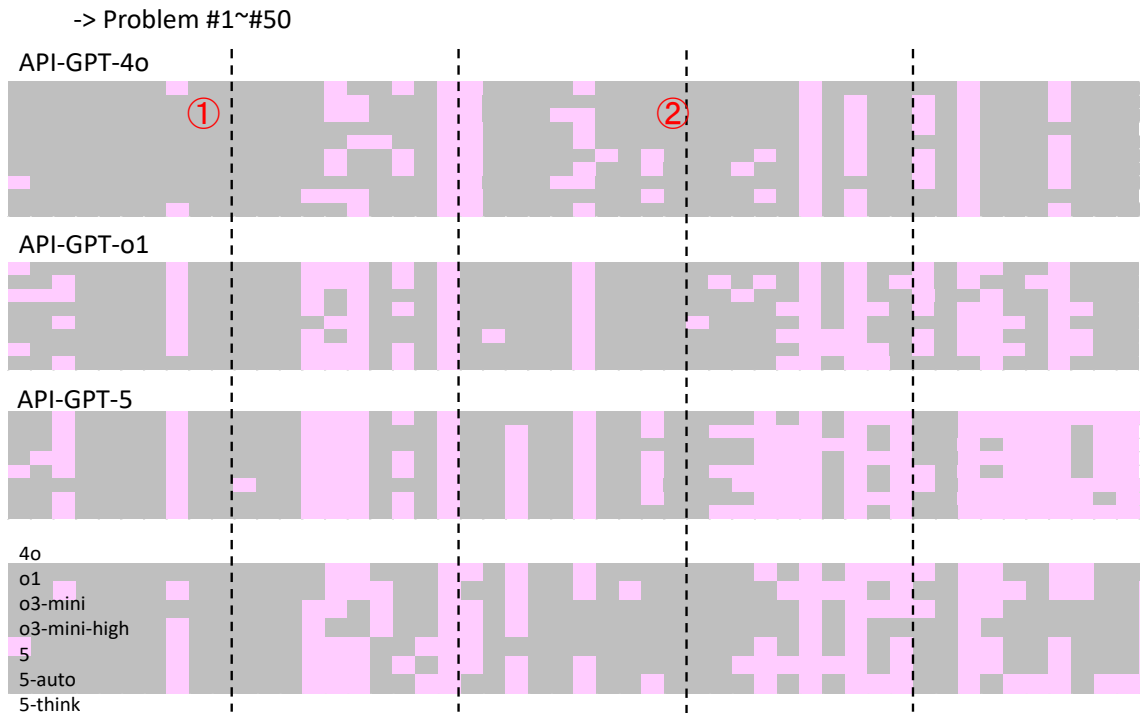

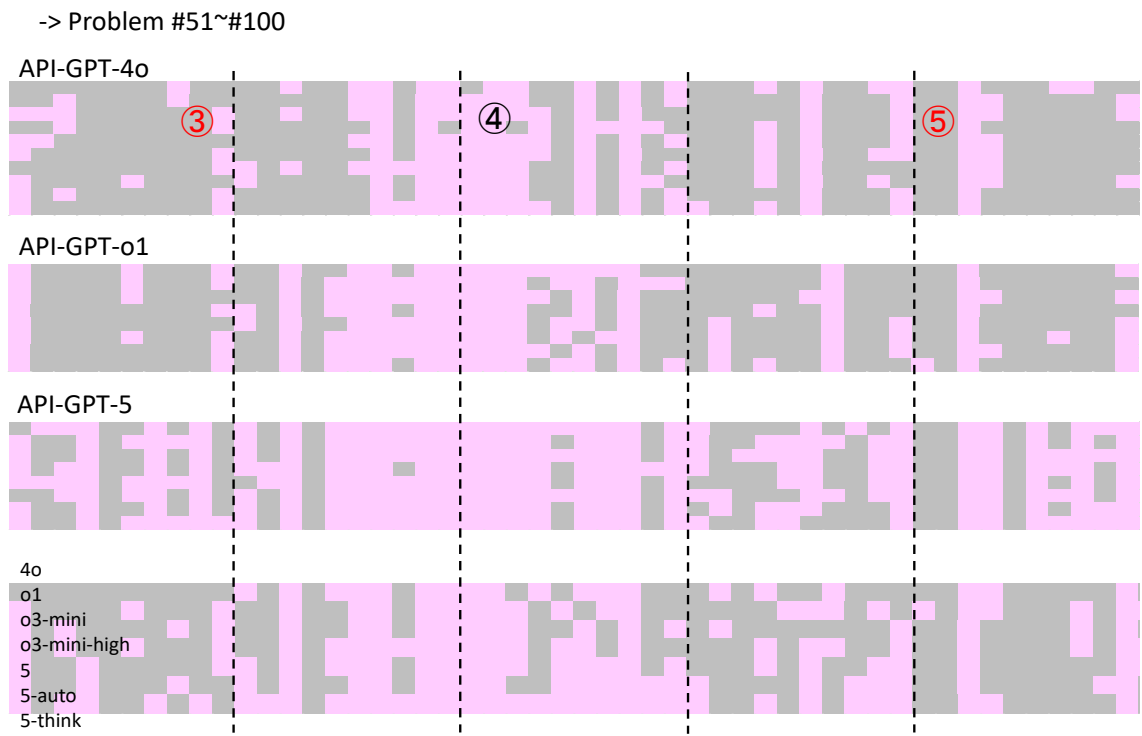

-> Problem #101~#137

API-GPT-4o ⑥ ⑦ ④

API-GPT-o1

API-GPT-5

4o
o1
o3-mini
o3-mini-high
5
5-auto
5-think

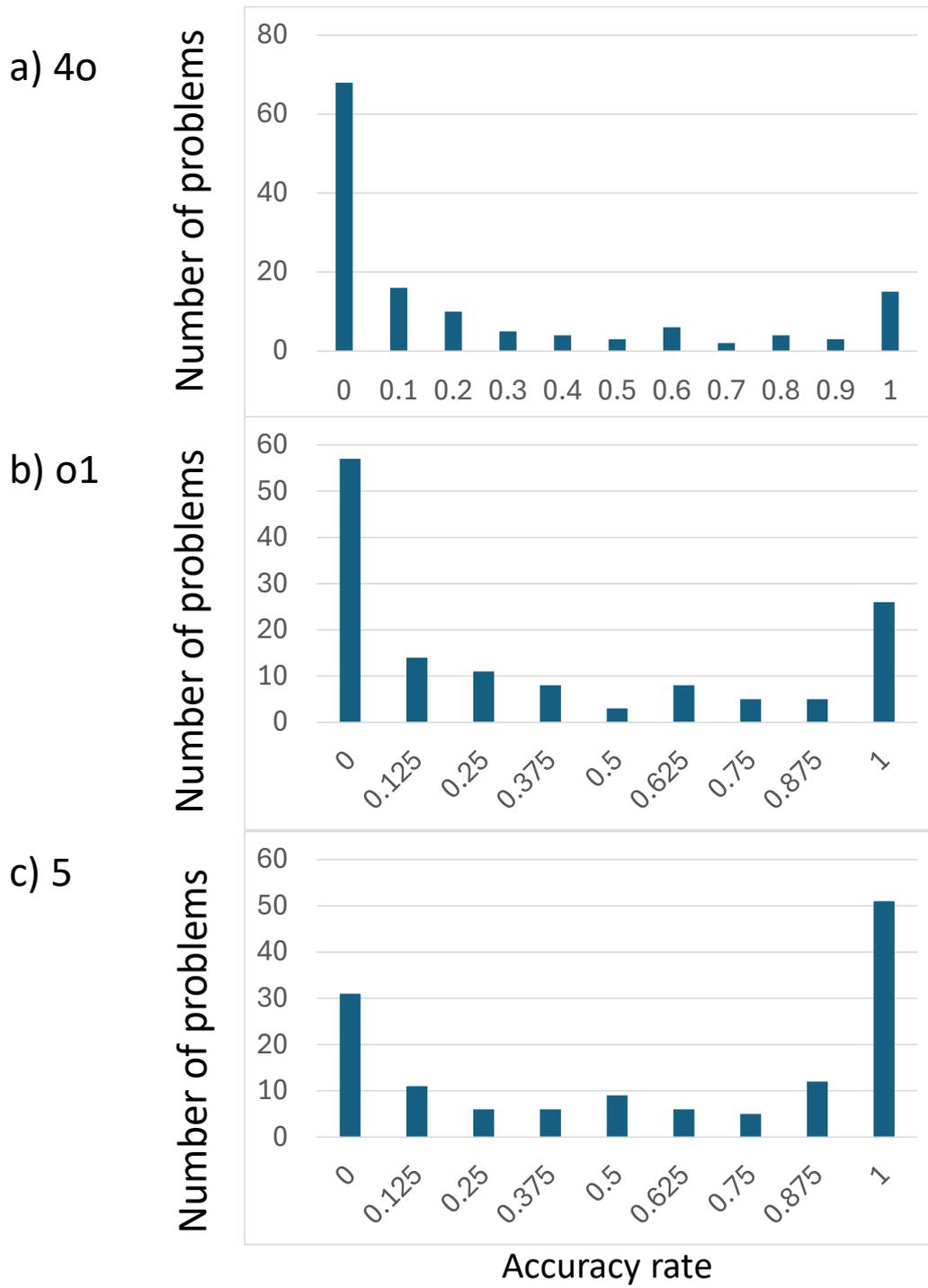

Figure 5. Distribution of accuracy rates for (a) GPT-4o, (b) GPT-o1, and (c) GPT-5.

Figure 6. Figures used to evaluate LLMs' ability to read numerical value texts embedded in figures. Solubility of carbon in ferrite phase and carbon composition at eutectic point 0.022 and 0.76 for (a), whereas they 0.0218 and 0.77 for (b).

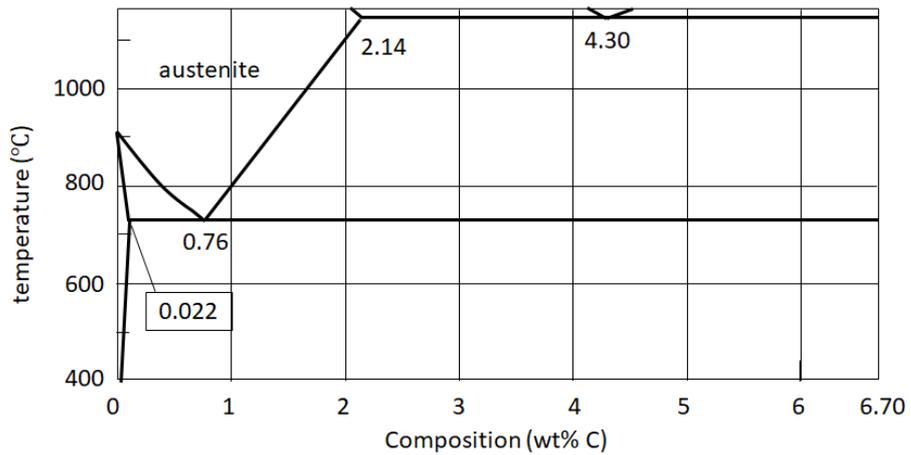

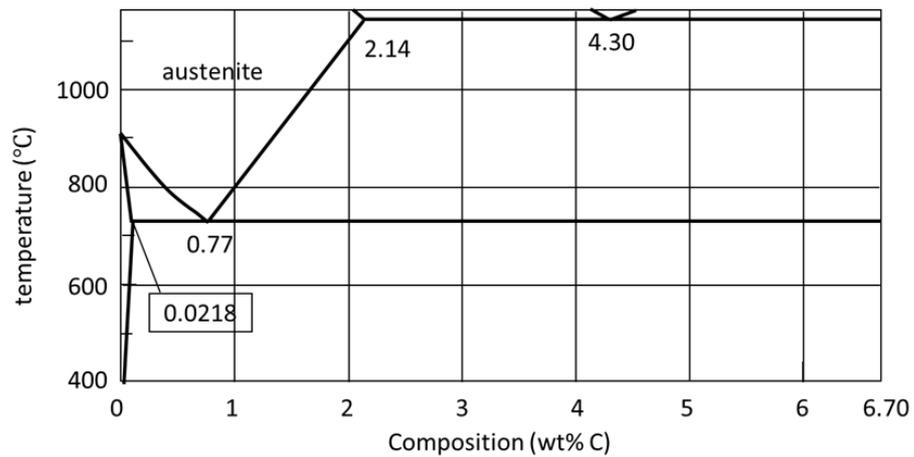